\documentclass[runningheads]{llncs}
\usepackage[T1]{fontenc}
\usepackage{graphicx}
\usepackage{multirow}
\usepackage{array}
\begin{document}
\title{Jumpstarting Surgical Computer Vision}
%
%
\author{Deepak Alapatt\inst{1} \and
Aditya Murali\inst{1} \and
Vinkle Srivastav\inst{1,2} \and
AI4SafeChole Consortium \inst{3,4,5,6} \and
Pietro Mascagni\inst{2,3} \and
Nicolas Padoy\inst{1,2}
}
%

\authorrunning{D. Alapatt et al.}
%
\institute{University of Strasbourg, CNRS, INSERM, ICube, UMR7357, Strasbourg, France \and
IHU-Strasbourg, Institute of Image-Guided Surgery, Strasbourg, France \and
Fondazione Policlinico Universitario Agostino Gemelli IRCCS, Rome, Italy \and
Azienda Ospedaliero-Universitaria Sant’Andrea, Rome, Italy \and
Fondazione IRCCS Ca’ Granda Ospedale Maggiore Policlinico di Milano, University of Milan, Milan, Italy
\and
Monaldi Hospital, AORN dei Colli, Naples, Italy
}
\titlerunning{Jumpstarting Surgical Computer Vision}
\maketitle              
\begin{abstract}
Consensus amongst researchers and industry points to a lack of large, representative annotated datasets as the biggest obstacle to progress in the field of surgical data science. Advances in Self-Supervised Learning (SSL) represent a solution, reducing the dependence on large labeled datasets by providing task-agnostic initializations. However, the robustness of current self-supervised learning methods to domain shifts remains unclear, limiting our understanding of its utility for leveraging diverse sources of surgical data. Shifting the focus from methods to data, we demonstrate that the downstream value of SSL-based initializations is intricately intertwined with the composition of pre-training datasets. These results underscore an important gap that needs to be filled as we scale self-supervised approaches toward building general-purpose ``foundation models'' that enable diverse use-cases within the surgical domain. Through several stages of controlled experimentation, we develop recommendations for pretraining dataset composition evidenced through over 300 experiments spanning 20 pre-training datasets, 9 surgical procedures, 7 centers (hospitals), 3 labeled-data settings, 3 downstream tasks, and multiple runs. Using the approaches here described, we outperform state-of-the-art pre-trainings on two public benchmarks for phase recognition: up to 2.2\% on Cholec80 and 5.1\% on AutoLaparo. 

\end{abstract}
\section{Introduction}
\label{sec1}
In the domain of surgery, the potential of artificial intelligence to disrupt clinical practice is becoming increasingly apparent with the development of models that can democratize expert-level decision-making \cite{mascagni2022artificial,madani2022artificial}, provide real-time intraoperative support \cite{mascagni2024early}, and generate video documentation \cite{mascagni2021computer} of procedures among various other applications. Still, a recent survey conducted by the Surgical Data Science Initiative \cite{maier2017surgical}, attributed a lack of tangible success stories to a lack of representative annotated data \cite{maier2022surgical}. Self-supervised learning (SSL) presents a promising approach to mitigating this reliance on large, well-annotated datasets by learning generic representations, and consequently, building generic model initializations. If sufficiently rich in information, these initializations could enable the development of foundation models for surgery, task-agnostic models that could robustly enable a diverse range of tasks with little to no labels. 

SSL methodology in the broader computer vision domain has seen significant developments in design, moving from heuristic-based pre-training tasks, such as predicting image augmentations, to more elegant strategies that directly learn to predict latent representations specific to the content of an image. Until recently, these shifts in design were slow to be adopted into the surgical domain with most work focusing on hand-crafted pre-text tasks, such as colorization \cite{ross2018exploiting}, or leveraging intrinsically collected information such as robot kinematics \cite{da2019self,sestini2021kinematic} or the temporal order of frames \cite{funke2018temporal,neimark2021train}. A large benchmark study of state-of-the-art SSL methodologies applied to the surgical domain attributed this to the complexity of translating and tuning newly proposed models to the surgical domain \cite{ramesh2023dissecting}. However, in the same study, the authors demonstrate through extensive experimentation that when applied correctly, these methods achieve state-of-the-art results on a range of tasks, benchmarks, and labeled-data regimes. Since then, several works have explored the value of pre-training similar methods on a massive scale by aggregating public datasets \cite{batic2024endovit}, utilizing large private repositories \cite{hirsch2023self}, or some combination of the two \cite{wang2023foundation}. All three of these works demonstrate that large-scale pre-training on their respective datasets brings sizable boosts in performance when performing various downstream tasks. \cite{batic2024endovit} aggregate 10 public datasets for several types of laparoscopic procedures sourced from centers worldwide to create a pre-training dataset of about 700k images. \cite{hirsch2023self} compile a private dataset of $\sim$7900 laparoscopic procedures ($>$23M images) from 8 centers and a curated polyp image dataset generated from $\sim$14k colonoscopies ($>$2M images). \cite{wang2023foundation} combine 6 public datasets for laparoscopy and endoscopy along with a large private repository of endoscopic videos to pre-train on $\sim$33k videos and $\sim$5M images. Different from these works that aggregate laparoscopic and endoscopic video data collected from different centers (hospitals) and different procedures to pre-train at scale, we focus this study on methodically clarifying how surgical pre-training datasets could and should be scaled to optimally leverage diverse surgical data. 

\begin{figure*}[]
\centering
\centerline{\includegraphics[width=1\linewidth]{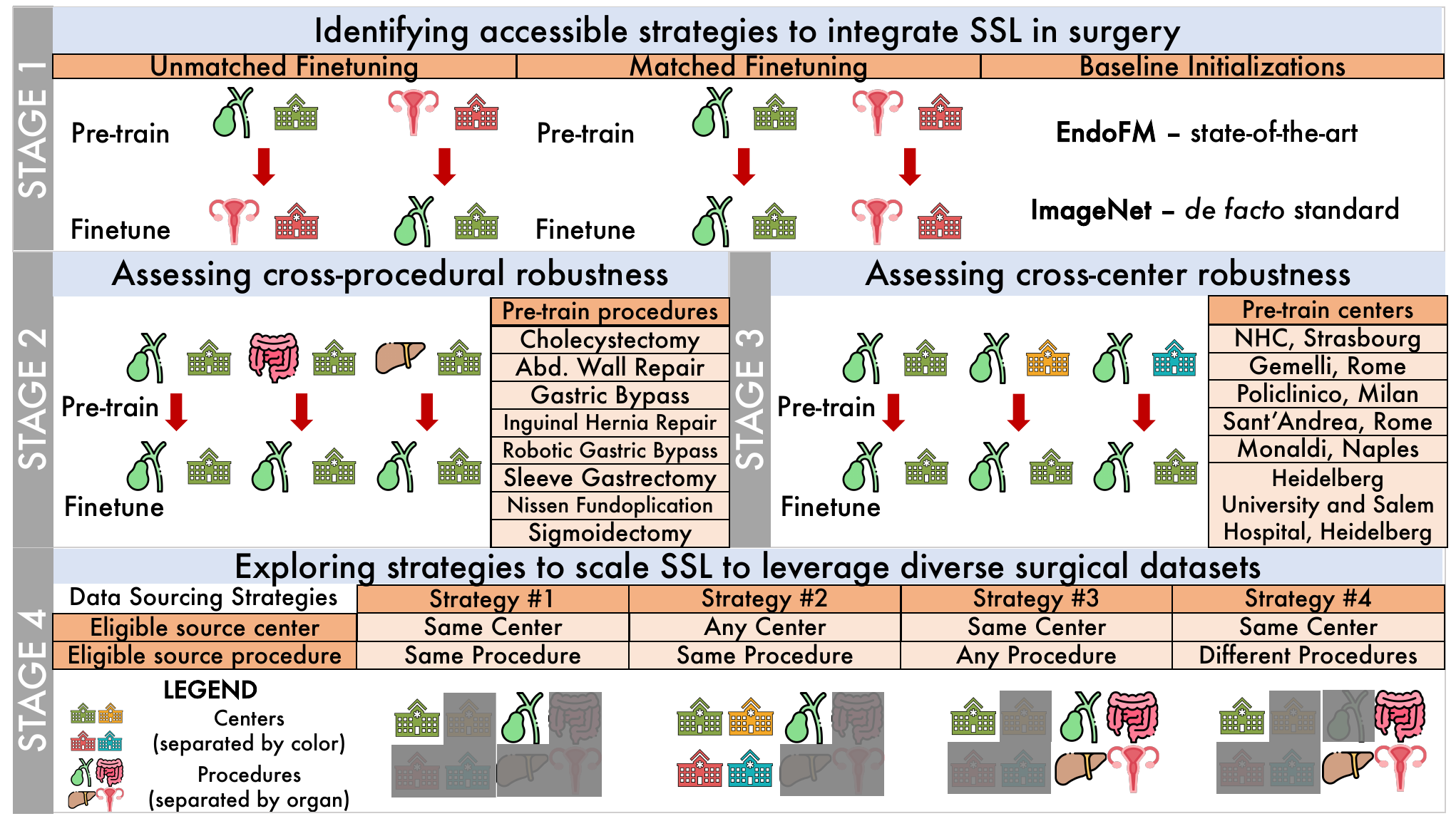}}
\caption{Overview of experimental design, aimed at addressing different objectives through several stages of study and over 300 experiments.}
\label{fig:umap}
\end{figure*}

Our contributions are threefold:
\begin{enumerate}
    \item We highlight important limitations when applying existing methodology, outperforming state-of-the-art approaches on two well-established public ben\-chmarks in surgical data science, Cholec80 and AutoLaparo.
   \
    \item We quantify the sensitivity of pre-training efficacy to various factors in over 300 experiments spanning 20 pre-training datasets, 9  procedures, 7 centers (hospitals), 3 labeled-data settings, 3 downstream tasks, and multiple runs.
    \item We validate our findings at scale ($\sim$400 videos) exploring different strategies to scale surgical SSL pre-training, quantifying key factors when doing so.
\end{enumerate}

\section{Methodology}\label{sec3}

In this section, we break down the various datasets and downstream tasks used to enable several distinct study objectives summarized in Figure \ref{fig:umap}.

\subsection{Datasets}

We use 7 different surgical computer vision datasets of minimally invasive abdominal procedures including:  AutoLaparo~\cite{wang2022autolaparo}, Cholec80~\cite{twinanda2016endonet}, 
    \linebreak Endoscapes-CVS201 ~\cite{murali2023endoscapes}, HeiChole~\cite{wagner2023comparative}, and Laparo425~\cite{kannan2019future}. We also use two private unlabeled Laparoscopic Cholecystectomy (LC) datasets: MultiChole2024 and StrasChole400, each containing $\sim$ 400 randomly selected cholecystectomy videos. StrasChole400 contains 400 videos from Strasbourg while MultiChole2024, an extension of MultiChole2022 \cite{kassem2022federated}, contains 405 videos, 200 from Strasbourg and $\sim$50 videos from each of the 4 centers (listed in Figure \ref{fig:umap}).

These datasets span 10 different surgical procedures (several cholecystectomy datasets, 7 others in Laparo420, 1 other in AutoLaparo), and 7 centers (several Strasbourg datasets, 4 others in MultiChole2024, 2 aggregated in HeiChole). 
Lastly, for fair experimentation, we remove 5 videos from Laparo425 that are found in the test sets for the downstream tasks, yielding Laparo420. 
Using these datasets, we create our various pre-training data combinations.

\subsection{Downstream Tasks}
We consider 3 downstream tasks, each using a different dataset: phase recognition (on both Cholec80 and AutoLaparo) and critical view of safety (CVS) prediction (Endoscapes-CVS201), thus spanning 2 procedures (cholecystectomy, hysterectomy) and 2 centers (Strasbourg, Hong Kong). Our selected downstream tasks are also highly diverse: CVS criteria assessment (3-class binary classification) is one of several tasks proposed for the prevention of bile duct injuries, requiring fine-grained analysis. 
Meanwhile, phase recognition is a widely studied and benchmarked task in surgical computer vision that, unlike CVS classification, requires long-range temporal reasoning. For each of the 2 phase recognition datasets, we consider one high-label data regime using 100\% of the available training labels (40 videos for Cholec80, 10 videos for Autolaparo) and two low-label data regimes using 5 videos and 3 videos, respectively. For the CVS task, we similarly consider 3 label settings using 120 videos (100\% of the train set), 30 videos (25\% of the train set), and 15 videos (12.5\% of the train set). For every low-label setting on each downstream task, we present results using the mean and standard deviation over 3 randomly sub-selected splits to mitigate the effects of selection bias. Following the bulk of previous works, we use F1-score to evaluate the phase recognition tasks and mean Average Precision (mAP) for CVS assessment.

\subsection{Training Procedure}

For all experiments, we use MoCo v2 \cite{chen2020improved} to pre-train a ResNet-50 feature extractor.
MoCo v2 is a contrastive learning method that learns powerful image representations by minimizing differences between embeddings of augmented views of the same image to be similar and maximizing distance between embeddings of different images.
We select MoCo v2 for all pre-training motivated by the results in~\cite{ramesh2023dissecting}, where it outperforms several competing methods across various surgical video analysis benchmarks. This pre-training procedure results in a trained ResNet-50 backbone, which we use to initialize a ResNet-50 classifier that we finetune separately for each downstream task while varying the number of labeled videos.
Finally, for the phase recognition downstream tasks (Cholec80, AutoLaparo), we further train TeCNO~\cite{czempiel2020tecno} on top of the finetuned ResNet-50 features to enable temporal reasoning as in~\cite{ramesh2023dissecting}. Detailed implementation settings are listed in the supplementary material.

\subsection{Study Stages}

\textbf{Stage 1.} 
The purpose of this stage is to establish performance baselines, exploring different application strategies of current SSL methodology to leverage related datasets (e.g. laparoscopic videos of related surgical procedures) for various tasks. To do this we perform 4 cross-over experiments examining the added value over ImageNet initializations: (1) where the downstream task is from the same dataset as the pre-training data (matched finetuning), and (2) where the downstream task and dataset are different from the pre-training dataset (unmatched finetuning). Note that both downstream tasks are conceptually similar (surgical phase recognition); however, the two datasets span two geographic regions (Hong Kong and France), two anatomical targets (Gallbladder and Uterus), and likely stylistic and functional differences in workflow and instrumentation. We adopt this process for two datasets, Cholec80 and AutoLaparo, reporting both matched and unmatched finetuning performance for both downstream phase recognition tasks. We only utilize public datasets in this stage to additionally illustrate the potential of utilizing accessible datasets. We also compare these strategies to EndoFM \cite{wang2023foundation}, the state-of-the-art massive-scale pretraining of a ViT-B model aimed at creating a foundation model for endoscopic data. 

\noindent\textbf{Stage 2.} In this stage, we aim to study the impact of pre-training surgical procedure types on downstream task performance. To enable this analysis, we separate Laparo420, which is a collection of various surgical videos, into 8 different subsets by procedure type (listed in Figure \ref{fig:umap}). We then evaluate on Cholec80 for phase recognition and Endoscapes-CVS201 for CVS Classification. In this controlled experiment, we keep the source center for all the pre-training datasets fixed (Strasbourg) and the pre-training dataset size fixed to $\sim$ 50 videos  \footnote{All stage 2 datasets contain exactly 50 videos except for cholecystectomy (45 videos), as 5 videos were removed to avoid contamination with the downstream task test sets.}; by controlling the source center and pre-training data scale in this fashion, we aim to isolate the impact of the pre-training procedure type. Still, confounding variables such as differences in instrumentation, workflow etc. do exist, which we aim to mitigate by including a large number of procedures (8) and large number of cases per procedure ($\sim$ 50).

\noindent\textbf{Stage 3.}\label{stage3} In the third stage of our study, we aim to study the impact of the source center for the pre-training dataset on downstream task performance. As has been noted by numerous works, surgical videos can vary greatly among medical centers due to differences in instrumentation, workflow, lighting, acquisition hardware, software, and patient characteristics. We leverage 4 subsets of MultiChole2024, which contains laparoscopic cholecystectomy videos from four different Italian hospitals, as well as the HeiChole public train set, combining cholecystectomy videos from 2 German hospitals. Further, we also utilize a 25-video subset of Cholec80; thus, constructing 6 subsets of $\sim$ 25 videos. As in stage 2, this allows us to control for procedure type (cholecystectomy) and data scale ($\sim$ 25 videos)\footnote{All center-specific pre-training datasets contain exactly 25 videos except for the public HeiChole training dataset, which contains 24 videos.} and isolate the impact of source center. Note again, confounding variables such as differences due to instrumentation, workflow, etc. do exist, which we aim to mitigate by including a large number of centers (2 from HeiChole and 5 others) and a large number of cases per center ($\sim$ 25). We again evaluate phase recognition and CVS classification.

\begin{table}[t]
\centering
\scalebox{0.9}{
\begin{tabular}{cccccc}
\multirow{1}{*}{Pre-training Dataset} & \multirow{2}{*}{\# Videos} & \multicolumn{2}{c}{Source Center} & \multicolumn{2}{c}{Procedure Type} \\
 & & Strasbourg & Other & Cholecystectomy & Other \\ \hline
StrasChole400 & 400 & 400 & 0 & 400 & 0 \\ 
MultiChole2024 & 405 & 200 & 205 & 405 & 0 \\
Laparo420 (with LC) & 420 & 420 & 0 & 45 & 375 \\
Laparo420 (w/o LC) & 375 & 375 & 0 & 0 & 375 \\
\end{tabular}}
\caption{Large Scale Pre-training Data Combinations}
\label{tab:stage4}
\end{table}

\noindent\textbf{Stage 4.} Lastly, one of the great promises of SSL is its scalability; however, there are numerous possibilities to accomplish this scaling in the surgical domain. To name a few, only including data from the same center and procedure type as the downstream task, fixing the center and varying procedure type or vice-versa, or naively including any and all available videos. 
To identify an optimal scaling strategy, we pre-train models on a series of data combinations, enumerated in Table \ref{tab:stage4}.
We again consider downstream Cholec80 phase recognition and CVS Classification as in Stages 2 and 3.

\section{Results and Discussion}\label{sec4}

\begin{table}[t]
\centering
  \resizebox{\textwidth}{!}{ 
\begin{tabular}{llllllll}
\multirow{3}{*}{Initialization} & \multirow{3}{*}{Arch.} & \multicolumn{3}{c}{Cholec80 Phase (F1)} & \multicolumn{3}{c}{AutoLaparo Phase (F1)} \\
                               & & \multicolumn{3}{c}{Number of Labeled Videos}              & \multicolumn{3}{c}{Number of Labeled Videos}                \\\hline
                                && \multicolumn{1}{c}{40}         & \multicolumn{1}{c}{5}                     & \multicolumn{1}{c}{3}                    & \multicolumn{1}{c}{10}          & \multicolumn{1}{c}{5}                     & \multicolumn{1}{c}{3}                     \\ \hline
Cholec80 SSL (Ours)      & R50              & 79.6       & \textbf{66.1 $\pm$ 4.2}       & \textbf{61.7 $\pm$ 4.8}       & 61.5        & 58.3 $\pm$ 2.8       & 47.7 $\pm$ 2.3       \\
AutoLaparo SSL (Ours)    & R50            & 80.0       & 62.4 $\pm$ 6.0       & 54.4 $\pm$ 5.7      & \textbf{70.7}        & \textbf{64.4 $\pm$ 5.1}       & \textbf{59.5 $\pm$ 3.9}       \\
EndoFM     & ViT-B            & 78.4       & 65.9 $\pm$ 3.9       & 59.5 $\pm$ 5.9      & 68.9        & 59.3 $\pm$ 5.0                      & 56.6 $\pm$ 5.2\\
\hline
ImageNet Supervised   & R50          & 80.3       & 62.3 $\pm$ 7.4       & 50.4 $\pm$ 3.8      & 70.1        & 57.6 $\pm$ 5.1                      & 50.8 $\pm$ 6.3   \\
ImageNet Supervised   & ViT-B          & \textbf{80.4}       & 60.1 $\pm$ 1.4       & 54.3 $\pm$ 4.4      & 68.6        & 62.3 $\pm$ 1.0                      & 58.5 $\pm$ 2.0
\end{tabular}
}
\caption{Stage 1 Results. Bold value represents the highest value for each column. The standard deviation is calculated over the 3 different selections of training data for the low-label settings.}
\label{tab:stage1_results}
\end{table}
\textbf{Stage 1: Cross-over Experiments.}
In this stage, we explore several accessible strategies to apply SSL-based pre-training initializations on two public benchmarks (Cholec80 and AutoLaparo) for phase recognition. In Table \ref{tab:stage1_results}, we see that for the matched pre-trainings, i.e. performing SSL pre-training on the same dataset as the downstream task, there are sizable boosts in performance; particularly so when many labeled examples aren't available for supervised finetuning. Quantitatively, when using as few as 3 labeled videos, we see an 11.3 \%  and 8.7 \% boost over the corresponding ImageNet initialization (R-50) when performing matched pre-training on Cholec80 and AutoLaparo, respectively. In contrast, other than the lowest labeled setting (3 labeled videos), the unmatched pre-trainings perform on par or worse than using corresponding ImageNet initializations (R-50) despite both being laparoscopic video datasets of abdominal procedures. We posit that SSL can help integrate critical representational attributes in low data settings that are otherwise easily learned when more data is available. Further, using the proposed matched finetuning approach on accessible training data (<90k public frames), we outperform EndoFM, the state-of-the-art foundation model for endoscopic data (trained on >5M frames, >70\% private) on two public endoscopic benchmarks by upto 2.2\% on Cholec80 and 5.1\% on AutoLaparo. This highlights the need to carefully consider the application strategy of SSL to both complement and contextualize methodological developments.

\begin{figure*}[t]
\centering
\centerline{\includegraphics[width=0.78\linewidth]{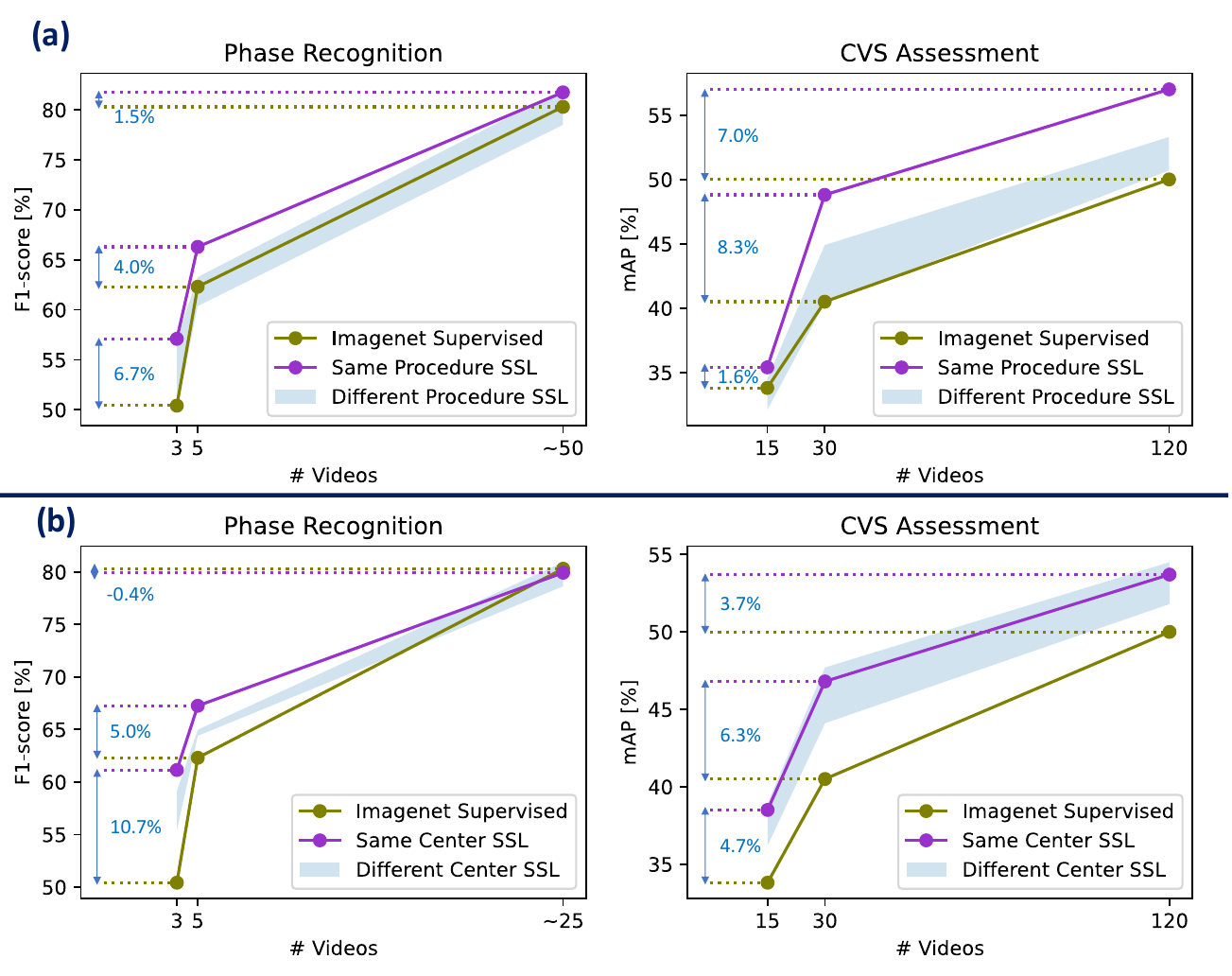}}
    \caption{Phase recognition and CVS assessment results for (a)Stage 2  and (b)Stage 3 . In Stage 2, SSL initializations are pre-trained on the relevant (violet line) or various different procedures as the downstream task (shaded blue area). The shaded area represents the range of performance across 7 non-cholecystectomy procedures.  In Stage 3, SSL initializations are pre-trained on the relevant (violet line) or various different centers as the downstream task (shaded blue area). The shaded area here represents the range of performance across 6 non-Strasbourg (2 aggregated in HeiChole) centers.}
\label{fig:procedure_label}
\end{figure*}

\begin{table}[]
\centering
\begin{tabular}{p{3.7cm}llllll}
\multirow{3}{*}{Initialization} & \multicolumn{3}{c}{Cholec80 Phase (F1)} & \multicolumn{3}{c}{Endoscapes-CVS201 (mAP)} \\
                                & \multicolumn{3}{c}{Number of Labeled Videos}              & \multicolumn{3}{c}{Number of Labeled Videos}                \\\hline
                                & \multicolumn{1}{c}{40}         & \multicolumn{1}{c}{5}                     & \multicolumn{1}{c}{3}                    & \multicolumn{1}{c}{120}          & \multicolumn{1}{c}{30}                     & \multicolumn{1}{c}{15}                     \\\hline
StrasChole400 SSL                    & 82.5       & 67.4 $\pm$ 4.9       & 62.2 $\pm$ 5.1       & 52.2        & 45.2 $\pm$ 4.7       & 35.0 $\pm$ 7.4       \\
MultiChole2024 SSL                  & 82.2       & 68.4 $\pm$ 1.3       & 61.1 $\pm$ 4.5      & 51.4        & 46.6 $\pm$ 3.5       & 34.2 $\pm$ 6.4       \\
Laparo420 (with LC) SSL                  & 79.1       & 63.9 $\pm$ 3.2       & 58.1 $\pm$ 6.3      & 53.3        & 44.1 $\pm$ 5.4       & 34.0 $\pm$ 6.6       \\
Laparo420 (w/o LC) SSL                  & 80.5       & 64.7 $\pm$ 3.3       & 55.7 $\pm$ 5.7      & 54.5        & 41.2 $\pm$ 4.5       & 31.5 $\pm$ 2.7       \\\hline
ImageNet Supervised             & 80.3       & 62.3 $\pm$ 7.4       & 50.4 $\pm$ 3.8      & 50.0        & 40.5 $\pm$ 7.0                      & 33.8 $\pm$ 4.1      
\end{tabular}
\caption{Stage 4 Results. The standard deviation is calculated over the 3 different selections of training data for the low-label settings.}
\label{tab:stage4_results}
\end{table}


\textbf{Stage 2: Cross-Procedure Experiments.} In Fig. \ref{fig:procedure_label}, we see that pre-training on the procedure relevant to the downstream tasks (cholecystectomy) brings boosts in performance over ImageNet initializations across the board for all labeled data settings and for both downstream tasks. Interestingly, using the 7 initializations other than those pre-trained on the same procedure as the downstream task delivers performance markedly worse, even lower than ImageNet initializations at times. This key result is both suggestive of the potential of accessible procedure-specific initializations and prompts important questions on how data from other procedures could be leveraged. In Stage 4, we further investigate whether merging subsets of data generated from procedures that are not directly relevant to the downstream task could bring additional value.


\textbf{Stage 3: Cross-Center Experiments.}
The stage 3 experiments (Fig. \ref{fig:procedure_label} (b)), present a number of interesting findings. First, similar to the previous stage, we see large boosts (up to 10.7 $\%$ for phase recognition, 7.2\% for CVS assessment) in performance over ImageNet initializations across all data settings and both tasks. Second, when pre-training, domain shifts introduced by sourcing videos from different centers appear to be much less impactful than those introduced by varying procedure types. This effect is resoundingly confirmed across the 7 centers (2 aggregated in HeiChole) included in this stage. Unlike the cross-procedure experiments, all initializations, irrespective of which center's data it was generated using, deliver boosts over the ImageNet initialization, at times even higher than initializations generated using data from the same center. These results suggest that having large-scale procedure-specific initializations such as those generated here, even from different centers, could enable rapid prototyping.

\textbf{Stage 4: Approaches to Scale.}
The previous 3 sections present strong evidence that pre-training dataset composition can severely impact downstream performance and must be carefully considered. In this last stage, our goal is to validate findings at scale, understand the interplay between different effects, and offer insights for optimal scaling of SSL in surgical computer vision. In Table \ref{tab:stage4_results}, when using initializations pre-trained on large ($\sim$ 400 videos) cholecystectomy datasets (StrasChole400 and MultiChole2024), we see performance largely on par with the cholecystectomy-only datasets included in previous study stages despite being an order of magnitude larger. An important caveat to this finding, though consistent with previous works \cite{ramesh2023dissecting}, is that higher capacity models may be better suited, if not required, to effectively leverage this scale of data for self-supervised pre-training \cite{hirsch2023self}. 
Finally, we aim to provide indications about the impact of curating multi-procedure datasets for SSL pre-training using two $\sim$400-video scale datasets, Laparo420 (with LC) and Laparo420 (w/o LC), containing 45/420 and 0/375 cholecystectomy procedures, respectively. Both datasets, despite being of comparable size with StrasChole400 and MultiCholec2024, perform markedly worse, reinforcing the findings of Stage 2 that learning procedure-agnostic initializations may require dedicated methodological design. To the best of our knowledge, each of the 3 previous works utilizing massive scale pre-training \cite{batic2024endovit,hirsch2023self,wang2023foundation} have only validated their pre-trained initializations on downstream tasks well-represented in their pre-training data. These findings further emphasize the need for more data-centric studies as we scale SSL methodology toward building a true procedure-agnostic foundation model for surgery.

\section{Conclusion}\label{sec5}

Recent developments in self-supervised learning for surgery allow us to learn generic representational features, useful for a range of downstream tasks in the domain. In this work, we build on these developments, shifting the focus from models to data, and explore whether pre-trained networks could be an effective means to consolidate disparate datasets in the field of Surgical Data Science. We try to illustrate, that simple model initializations could help promote work on new problems in new procedures at new clinical centers with fewer annotations. The pre-trained initializations generated through this work will be made available alongside the released code at \url{https://github.com/CAMMA-public/ScalingSurgicalSSL/}.

\section{Acknowledgement}
This work was supported by French state funds managed by the ANR under Grant ANR-20-CHIA-0029-01 (Chair AI4ORSafety) and Grant ANR-10-IAHU-02 (IHU Strasbourg), and by BPI France under reference DOS0180017/00 (project 5G-OR). The AI4SafeChole consortium is represented by Ludovica Baldari, Giovanni Guglielmo Laracca, Ludovica Guerriero, Segio Alfieri, Claudio Fiorillo, Giuseppe Quero, Elisa Cassinotti, Luigi Boni, Diego Cuccurullo, Guido Costamagna, and Bernard Dallemagne.
\subsubsection{\discintname}
The authors have no competing interests to declare. 
%
%
%
%
\bibliographystyle{splncs04}
\bibliography{sn-bibliography1}
\end{document}